\documentclass{article}

    \PassOptionsToPackage{numbers, compress}{natbib}

    \usepackage[preprint]{neurips_2019}

\usepackage[utf8]{inputenc} 
\usepackage[T1]{fontenc}    
\usepackage{hyperref}       
\usepackage{url}            
\usepackage{booktabs}       
\usepackage{amsmath, amssymb, amsfonts}       
\usepackage{nicefrac}       
\usepackage{microtype}      
\usepackage{graphicx, float, subcaption}
\usepackage{amsmath}
\DeclareMathOperator*{\argmax}{arg\,max}

\title{Autoregressive Models: What Are They Good For?}

\author{%
  Murtaza Dalal\thanks{Authors contributed equally to this work, ordered alphabetically by surname.} \\
  \texttt{mdalal@berkeley.edu} \\
  \And
  Alexander C. Li\footnotemark[1] \\
  \texttt{alexli1@berkeley.edu} \\
  \And
  Rohan Taori\footnotemark[1] \\
  \texttt{rohantaori@berkeley.edu} \\
}

\begin{document}

\maketitle
\vspace{-0.5cm}
\begin{abstract}
Autoregressive (AR) models have become a popular tool for unsupervised learning, achieving state-of-the-art log likelihood estimates. We investigate the use of AR models as density estimators in two settings -- as a learning signal for image translation, and as an outlier detector -- and find that these density estimates are much less reliable than previously thought. We examine the underlying optimization issues from both an empirical and theoretical perspective, and provide a toy example that illustrates the problem. Overwhelmingly, we find that density estimates do not correlate with perceptual quality and are unhelpful for downstream tasks.
\end{abstract}

\section{Introduction}
Autoregressive (AR) models are a class of likelihood models that attempt to model the data distribution by estimating the data density. They approach the maximum likelihood objective $\theta^* = \argmax_{\theta} \mathbb{E}_{x \text{\textasciitilde{}} p_{data}(x)} [\log p_{\theta}(x)]$ by factorizing $p_{\theta}(x)$ over the dimensions of $x$ via the chain rule. They learn each conditional probability forming $p_{\theta}(x) = \Pi_{i=1}^d p_\theta(x_i|x_1, ..., x_{i-1})$, and this decomposition helps them achieve negative log-likehood (NLL) scores superior to other methods such as VAEs \cite{Kingma2013Auto-EncodingBayes} or flow models \cite{Dinh2016DensityNVP,Bach2007Glow:Diederik}. 

PixelCNN \cite{Courville2016PixelNetworks} was the first to introduce a convolutional AR architecture, achieving a NLL of 3.00 bits/dim on CIFAR-10. Over the past few years, a flurry of further modifications  \cite{Oord2016ConditionalDecoders,Parmar2018ImageTranformer,Salimans2017PixelCNN++:Modifications,GrayGPUWeights,Chen2017PixelSNAIL:Model} has pushed the score down to 2.85 bits/dim, the best known reported NLL on CIFAR-10 to date. However, despite these advancements, their uses outside of compression have not been well explored. Samples from these models are considerably worse than state-of-the-art GANs \cite{Brock2018LargeSynthesis}, and they do not provide a compact latent space representation, an important piece for use in downstream tasks. 

Prior work \cite{Theis2015AModels} has shown that AR metrics such as log-likelihood and Parzen window estimates are poor indicators of the AR model's performance on specific tasks. We investigate this empirically in two scenarios: using the log-density as a learning signal for image translation, and for outlier detection. Our results show that density estimates neither correlate with perceptual quality, nor are useful for downstream tasks. 

\section{Log-Likelihood Estimates as Learning Signal} \label{sec:arcgan}

We first explore using NLL scores for image-to-image translation.  CycleGAN \cite{Zhu2017UnpairedNetworks}, a popular unpaired image translation method, learns mappings between domains $X$ and $Y$ using GANs \cite{Goodfellow2014GenerativeNetworks}. However, GANs are known to be unstable during training due to their adversarial framework, and they also lack an evaluation metric for the perceptual quality of generated images, so there is no way to evaluate the learned mapping other than visually inspecting the samples. 

Replacing the CycleGAN discriminator with NLL estimates from an AR model seems like a natural solution to these problems. Powerful AR models provide log-likelihood estimates of our samples throughout training that we can compare across methods, and optimization could be easier since there is no longer an adversarial minimax problem, which could help learn more general cross-domain mappings.

\subsection{ARCycle Formulation}
For the mapping functions $F: X \rightarrow Y$ and $G: Y \rightarrow X$, we borrow the same cycle consistency loss from \cite{Zhu2017UnpairedNetworks}: 
\begin{equation}
    \mathcal{L}_{cyc}(G, F) = \mathbb{E}_{y \sim p_{data}(y)}[||F(G(y))-y||_1] + \mathbb{E}_{x \sim p_{data}(x)}[||G(F(x))-x||_1]
\end{equation}
Instead of an adversarial loss, we use a generative loss with the negative log-likelihood of the generated image under our autoregressive model. This autoregressive model is trained purely on real images from its domain. For the mapping function $F: X \rightarrow Y$ and the density model $P$ on $Y$, we express the objective as:
\begin{equation}
    \mathcal{L}_{NLL}(P, F, X, Y) = \mathbb{E}_{x \sim p(x)}[-\log P(F(x))]
\end{equation}
Thus, our overall ARCycle objective that we minimize is 
\begin{equation}
    \mathcal{L}(F, G, P_X, P_Y) = \mathcal{L}_{NLL}(P_Y, F, X, Y) + \mathcal{L}_{NLL}(P_X, G, Y, X) + \beta \mathcal{L}_{cyc}(G, F)
    \label{eq:arcgan}
\end{equation}

\subsection{Experiments and Discussion}
\begin{figure}[h]
    \centering
    \begin{subfigure}[t]{0.18\textwidth}
        \centering
        \includegraphics[width=\textwidth]{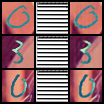}
        \caption{$\mathcal{L}_{NLL} + \mathcal{L}_{cyc}$}
        \label{fig:1a}
    \end{subfigure}%
    ~ 
    \begin{subfigure}[t]{0.18\textwidth}
        \centering
        \includegraphics[width=\textwidth]{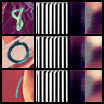}
        \caption{Just $\mathcal{L}_{NLL}$}
        \label{fig:1b}
    \end{subfigure}%
    ~ 
    \begin{subfigure}[t]{0.18\textwidth}
        \centering
        \includegraphics[width=\textwidth]{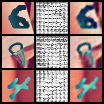}
        \caption{Just $\mathcal{L}_{cyc}$}
    \end{subfigure}
    ~ 
    \begin{subfigure}[t]{0.18\textwidth}
        \centering
        \includegraphics[width=\textwidth]{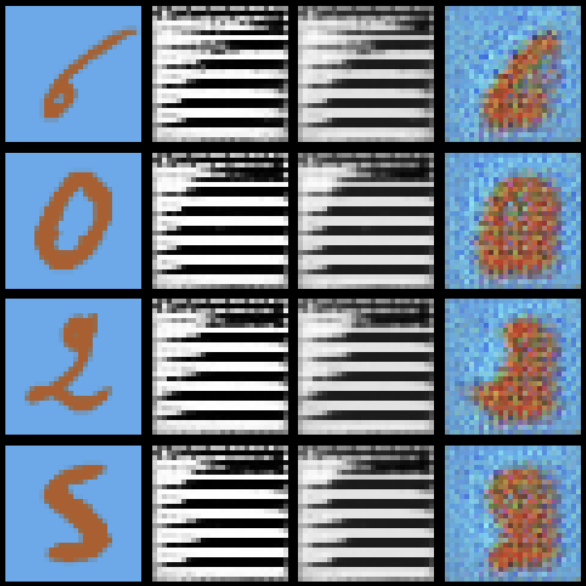}
        \caption{$\mathcal{L}_{NLL} + \mathcal{L}_{cyc}$, Gaussian blur on generator output.}
        \label{fig:1d}
    \end{subfigure}
    ~
    \begin{subfigure}[t]{0.18\textwidth}
        \centering
        \includegraphics[width=\textwidth]{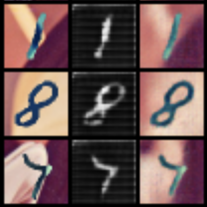}
        \caption{ARCycle starting with pretrained generators.}
    \end{subfigure}
    ~
    \caption{ARCycle trained in different settings. The left columns are real images from colored-MNIST dataset, the middle columns contain mappings of the images to the MNIST domain, and the right columns show reconstruction with the reverse mapping back to the colored-MNIST domain.}
    \label{fig:fig1}
\end{figure}

\begin{figure}[h]
    \centering
    \begin{subfigure}[t]{0.18\textwidth}
        \centering
        \includegraphics[width=\textwidth]{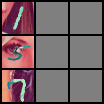}
        \caption{Iteration 0}
    \end{subfigure}%
    ~ 
    \begin{subfigure}[t]{0.18\textwidth}
        \centering
        \includegraphics[width=\textwidth]{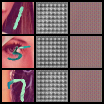}
        \caption{Iteration 25}
    \end{subfigure}
    ~
    \begin{subfigure}[t]{0.18\textwidth}
        \centering
        \includegraphics[width=\textwidth]{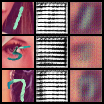}
        \caption{Iteration 50}
    \end{subfigure}
    ~
    \begin{subfigure}[t]{0.18\textwidth}
        \centering
        \includegraphics[width=\textwidth]{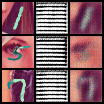}
        \caption{Iteration 75}
    \end{subfigure}
    ~
    \begin{subfigure}[t]{0.18\textwidth}
        \centering
        \includegraphics[width=\textwidth]{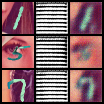}
        \caption{Iteration 100}
    \end{subfigure}
    \caption{AR Cycle quickly learns to output lines in the MNIST space when optimizing $\mathcal{L}_{NLL} + \mathcal{L}_{cyc}$. }
    \label{fig:figure_2}
\end{figure}

We trained ARCycle with the loss in \autoref{eq:arcgan}, as well as several ablations. We used a pre-trained PixelCNN++  to compute $\mathcal{L}_{NLL}$ and set $\beta$ so that $\mathcal{L}_{NLL}$ is on the same order as the reconstruction loss. After training to convergence, we observed the results in \autoref{fig:fig1}. 

The reconstructions in \autoref{fig:1a} are perfect, but the translated images in the MNIST domain have collapsed to a degenerate solution. The network encodes enough information into the translation to perfectly reconstruct the original image, but hits a local optimum for $\mathcal{L}_{NLL}$ and does not learn to produce realistic MNIST digits. The negative log prob of the translations starts at about $8$ bits/dim and steadily goes down to $3$ bits/dim, far from the $.8$ bits/dim that PixelCNN++ achieves on the MNIST test set. Even if we remove competing losses and only optimize $\mathcal{L}_{NLL}$ as in \autoref{fig:1b}, or blur the transformed images to remove high-frequency patterns as in \autoref{fig:1d}, our mappings still fail to produce realistic MNIST digits. Most interestingly, even if we initialize training with mappings $F$ and $G$ pretrained using CycleGAN, the ARCycle training procedure manages to corrupt the mappings by producing faint lines in the background. 


To analyze how the network learns to arrive at the degenerate solution, we plot reconstructions over iterations in \autoref{fig:figure_2}. Here, we see an interesting phenomenon: with more and more updates, the translated images look like a set of lines, while at the same time the reconstructions get more accurate. A hypothesis is that the network is learning to embed information in high-frequency signals that aren't apparent to the human eye, a phenomenon also mentioned in \cite{Zhu2017UnpairedNetworks}. 

$\mathcal{L}_{NLL}$ proved to be difficult to optimize in a variety of settings, and these results suggest that using AR density estimates as learning signal for optimization may be flawed, which we investigate in the next section.

\section{Optimization With AR Models}

\begin{figure}[h]
    \centering
    \begin{subfigure}[b]{0.55\textwidth} 
        \centering
        \includegraphics[width=\textwidth]{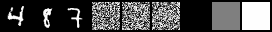}
        \includegraphics[width=\textwidth]{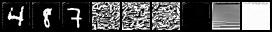}
        \includegraphics[width=\textwidth]{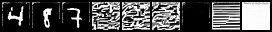}
        \caption{Optimized samples at iteration 0, 5000, and 10000. }
        \label{subfig:images}
    \end{subfigure}%
    ~ 
    \begin{subfigure}[b]{0.25\textwidth} 
        \centering
        \includegraphics[width=\linewidth]{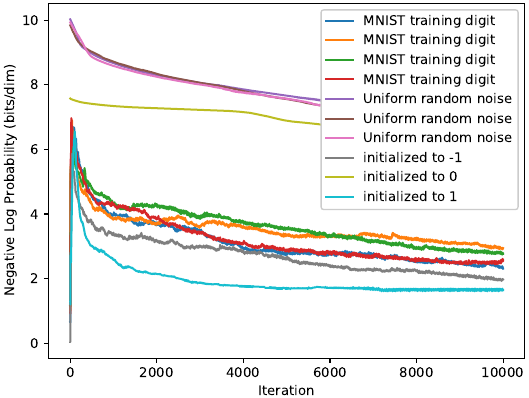}
        \caption{Loss curve over 10k iterations. }
        \label{subfig:directoptloss}
    \end{subfigure}
    \caption{Result of directly maximizing images' PixelCNN++ log-likelihood estimate by applying gradients to the image pixels themselves. }
    \label{fig: directopt}
\end{figure}

In this section, we investigate the optimization process and explain why an AR log-likelihood loss can be difficult to optimize in general.

\subsection{Directly Maximizing Image Log-Likelihood}
One key observation is that the gradient of the ARCycle loss with respect to our generator $F$'s parameters is entirely dependent -- by the gradient chain rule -- on the gradient of the log-likelihood with respect to the generated image. If that pixel-level gradient vanishes, gradient descent cannot help our generator produce better images.
Thus, we want to see if it's possible to directly optimize the image pixels themselves using gradient descent on the negative log-likelihood output of PixelCNN++. 

\autoref{fig: directopt} shows the result of applying gradient descent on 3 MNIST digits from the test set, 3 images of random noise, and black, gray, and white images. Every image begins to accumulate noise, even the true MNIST digits, which should have already been in a local minimum. Both the noisy and gray images form lines, which indicates that PixelCNN++ gradients lead us to images that locally resemble digits, rather than images that gradually form digits as training progresses. Not only does the model encourage local texture over global structure, but the optimization problem seems ill-conditioned. The true MNIST digits start with 0.6 bits/dim, but jump up to 7 bits/dim after a few gradient steps. 
Since the digits remain visually identical, log-likelihood has no bearing on the quality of a sample. Overall, minimizing log-likelihood is difficult to do and is not guaranteed to produce good results.


\subsection{Harder Optimization Problem}
\begin{figure}[h]
    \centering
    \begin{subfigure}[t]{0.25\textwidth}
        \centering
        \includegraphics[width=\linewidth]{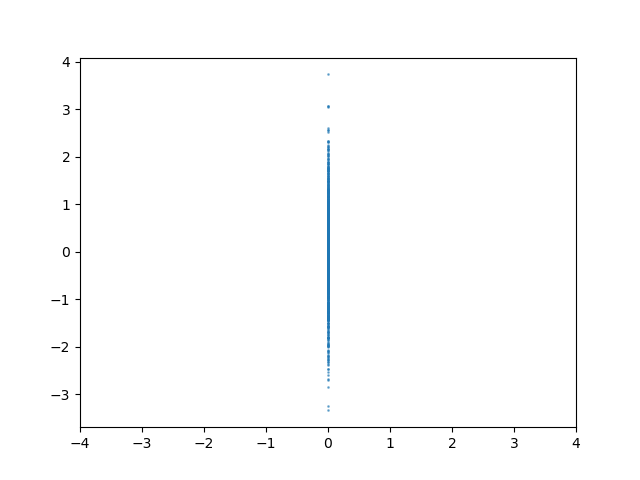}
        \caption{True data distribution.}
        \label{subfig:truedist}
    \end{subfigure}%
    ~ 
    \begin{subfigure}[t]{0.25\textwidth}
        \centering
        \includegraphics[width=\linewidth]{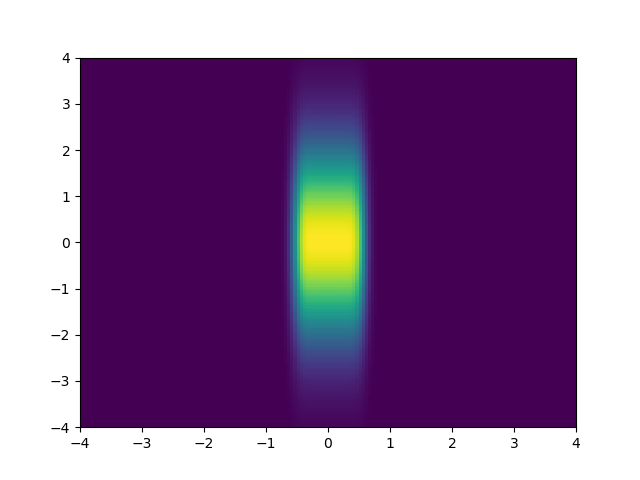}
        \caption{Learned probability with discretized Gaussian. }
        \label{subfig:learneddist}
    \end{subfigure}
    ~
    \begin{subfigure}[t]{0.25\textwidth}
        \centering
        \includegraphics[width=\linewidth]{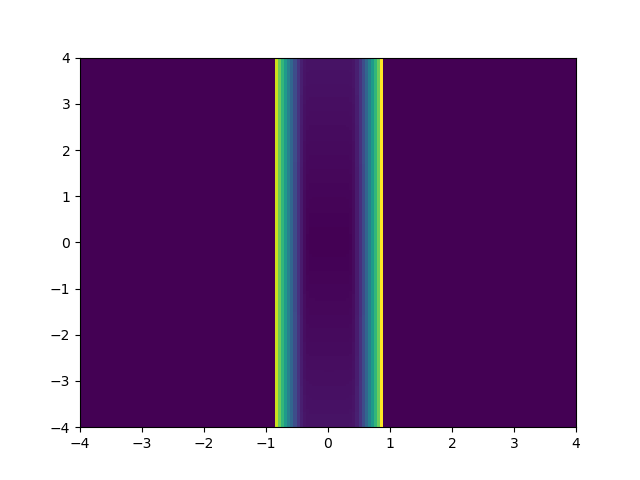}
        \caption{Gradient norm heatmap.}
        \label{subfig:gradient}
    \end{subfigure}
    \caption{2D toy example illustrating the optimization problem that arises when the data lies on a lower-dimensional manifold. }
    \label{fig:toy2d}
\end{figure}

Training PixelCNN++ on CIFAR-10 or ImageNet achieves good performance for a wide range of hyperparameter choices. Why, then, is it so hard to maximize the log-likelihood of samples under a trained PixelCNN++ model?

The PixelCNN++ objective is $\min_{\theta} \mathbb{E}_{x \sim P_{data}}[-\log P_{\theta}(x)]$.
As long as $P_\theta$ is initialized with support almost everywhere and the parameters $\theta$ are closely tied to the density estimate, the gradient of this is nonzero and should yield improvements to the AR model's ability to model the distribution. 

In contrast, by trying to directly optimize our samples, our objective is $\min_{P_{gen}} \mathbb{E}_{x \sim P_{gen}}[-\log P_{\theta}(x)]$ Instead of trying to increase our probability of data points by adapting our model, we have to find samples that have high likelihood under a fixed AR model. In certain scenarios, this proves to be an impossible task. Assuming that the data lies on a lower-dimensional manifold within the pixel-space, a powerful enough AR model will put all of its probability mass on the manifold, leaving none elsewhere. Thus, as the model has probability zero almost everywhere, the gradients are also zero almost everywhere, so gradient descent cannot improve the log-likelihood of points.

We visualize this with a 2-dimensional toy problem in \autoref{fig:toy2d}. The true data distribution, shown in \autoref{subfig:truedist}, always has $x_1 = 0$, while $x_2 \sim \mathcal{N}(0, 1)$. We use a MADE AR model \cite{Germain2015MADE:Estimation} to learn a discretized Gaussian log-likelihood, an analogue of the discretized mixture of logistics used by PixelCNN++. As seen in \autoref{subfig:learneddist}, the trained model is able to fit the true distribution fairly closely. Internally, the Gaussian distribution in the $x_1$-direction is becoming narrower, as it tries to place as much mass as possible in the bin for $x_1 = 0$.
This leads to the gradient heatmap in \autoref{subfig:gradient}, which is taken with respect to $x_1$ and $x_2$. The gradient is nonzero in only 2 thin strips, which will converge to two infinitely thin lines as the AR model fits the true distribution better and better. Optimizing a set of samples on this landscape would be impossible, as only the points on those thin, bright gradient strips would have a nonzero gradient. Thus, the ability of powerful autoregressive models to represent any distribution hurts downstream optimization by creating vanishing gradients almost everywhere. 

\section{Correlation between Log-Likelihood and Perceptual Features}
\subsection{Do Realistic Images Have High Log-Likelihood?}
\begin{figure}[h]
    \centering
    \vspace{-0.5cm}
    \begin{subfigure}[t]{0.55\textwidth}
        \centering
        \includegraphics[width=\linewidth]{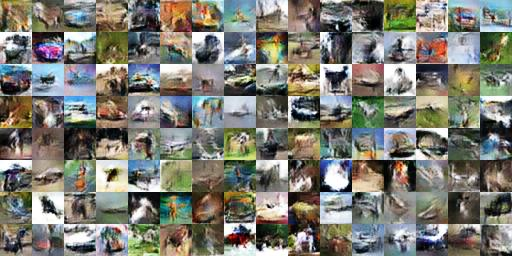}
    \end{subfigure}
    ~ 
    \begin{subfigure}[t]{0.4\textwidth}
        \centering
        \includegraphics[width=\linewidth]{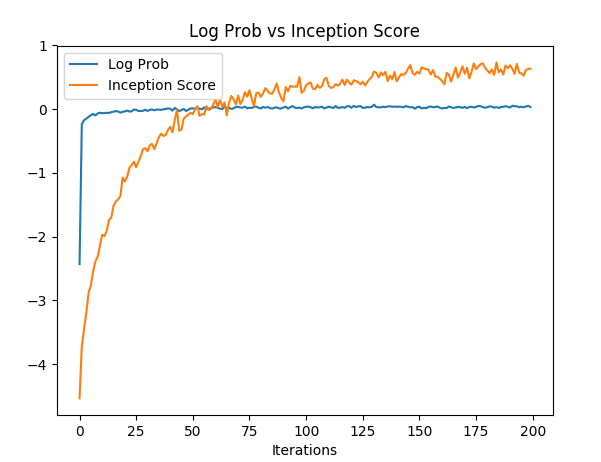}
        \centering
    \end{subfigure}
    ~
    \caption{Left: samples from WGAN-GP trained on CIFAR-10. Right: plot of the zero-centered inception score and PixelCNN++ negative log probability of WGAN-GP samples over the WGAN training process.
    }
    \label{fig:gan-samples-log-prob-vs-inception}
\end{figure}

Do images that look like the training dataset have high log probability under an AR model for that dataset? We propose the following experiment: 
Train a WGAN-GP \cite{Gulrajani2017ImprovedGANs} model to produce highly realistic CIFAR-10 images, compute the bits/dim of those samples under a PixelCNN++, and compare that quantity to the bits/dim of the CIFAR-10 test set under the same PixelCNN++.

Our WGAN-GP samples, which are visually indistinguishable from real images, obtained 6.52 bits/dim, significantly higher than the CIFAR-10 test set, which had 2.92 bits/dim. This suggests that perceptual similarity may not be correlated well with the AR log probability. To further investigate, we compare the AR log probability to the inception score, a metric purported to correlate well with human perception \cite{Salimans2016ImprovedDCGAN}. \autoref{fig:gan-samples-log-prob-vs-inception} shows that the log probability remains relatively constant, even though the inception score and sample perceptual quality progressively increase. This further implies that log probability is not a good metric to evaluate whether a given image is similar to a certain distribution of images. 

\subsection{Do Images with High Log-Likelihood Look Realistic?}

We seek to better understand whether images that have high log probability under an AR model trained on a dataset look like images from that dataset. To test this, we evaluated how well AR models fare against other methods of outlier detection for high dimensional data, where using simple strategies such as z-scores is not particularly useful. All the methods we test only assume knowledge of the CIFAR-10 training set and attempted to classify new images as CIFAR-10.

\begin{table*}
\centering
\begin{tabular}{lrrrr}
\toprule
\textbf{Dataset} & \textbf{AR-2SD} & \textbf{AR-1SD} & \textbf{AR-One-sided} & \textbf{CCG}\\
\hline
CIFAR-10 Test & 92.5\% & 68\% & 94.7\% & 92.3\% \\
WGAN-GP CIFAR-10 Samples & 0\% & 44.9\% & 0\% & 69.7\% \\
SVHN Test & 90.9\% & 44.9\% & 100\% & 37.3\% \\
Noise & 0\% & 0\% & 0\% & 1.1\% \\
All Black & 0\% & 0\% & 100\% & 0\% \\
All White & 0\% & 0\% & 100\% & 0\%\\
\bottomrule
\end{tabular}
\caption{The percent of the dataset classified to be in CIFAR-10, i.e. classified as \textbf{not} an outlier.}
\label{table:outlier-exps}
\vspace{-0.5cm}
\end{table*}

We tried three strategies for outlier detection with AR models. We computed the mean $\mu$ and standard deviation $\sigma$ bits/dim over the CIFAR-10 training set using the PixelCNN++'s NLL. We then constructed 3 intervals of the following form: 2 Standard Deviation Interval (AR-2SD): $[\mu - 2\sigma, \mu + 2\sigma]$, 1 Standard Deviation Interval (AR-1SD): $[\mu - \sigma, \mu + \sigma]$, and One-Sided Interval (AR-One-sided): $(-\infty, \mu + 2\sigma]$.
We classify an image as an outlier if its NLL in bits/dim lies outside the interval. 

We compare against a Class-Conditional Gaussians (CCG), a method for high-dimensional outlier detection using deep classifiers \cite{Lee2018AAttacks}.
CCG is as follows: Train a classifier on the dataset (Inception-v1 \cite{Szegedy2015GoingConvolutions} on CIFAR-10), strip off the output layer, and fit class conditional Gaussians to the feature vectors given by evaluating the classifier on the dataset. For a new test image, compute its feature vector and evaluate the feature vector's probability under each class conditional Gaussian; if the probability is less than some threshold, classify as an outlier. 

 We test the effectiveness of the outlier detection methods on six different datasets: CIFAR-10 test set, samples from WGAN-GP trained on CIFAR-10, SVHN test set, random noise images from $\{0, ..., 255\}^D$, completely black images, and completely white images. Intuitively, we would expect a good outlier detector to classify CIFAR-10 test set images correctly as CIFAR, WGAN-GP samples as CIFAR most of the time, and SVHN, noise, all black and all white images as outliers. 
 
 \autoref{table:outlier-exps} shows that CCG largely does what a good outlier detector should do, but the AR model does not. In fact, the AR model often classifies SVHN to be CIFAR just as often as it does actual CIFAR-10 images. As found earlier by \cite{Nalisnick2018DoKnow}, PixelCNN++ actually assigns higher log probability to images from SVHN than actual CIFAR-10 images -- SVHN achieves 2.1 bits/dim, while CIFAR-10 only gets 2.92 bits/dim. Additionally, the model assigns even high log probability to all black and all white images: $0.008$ and $0.16$ bits/dim respectively. These failures on outlier detection indicate that log probability under an AR model is not usefully correlated with our notion of perceptual quality.

\section{Conclusions and Future Work}
We investigate the usefulness of the density estimates of autoregressive models. We apply them to 2 tasks: as a learning signal for unsupervised image translation with ARCycle, and outlier detection, and find that the density estimates are not informative in both settings. We also perform an analysis on the underlying optimization issues, finding that optimizing using AR models leads to degenerate solutions due to vanishing gradients. Their log-likelihood estimates also don't correlate with perceptual quality, which explains their poor performance at outlier detection. Given that these models achieve superior log-likelihood, our findings call into question the overall utility of likelihood-based learning. In this work, we examined PixelCNN++ in detail; one interesting avenue of future work is investigating if our results hold for other autoregressive models as well.

\nocite{*}
\bibliographystyle{plainnat}
\bibliography{references}

\end{document}